\documentclass[twocolumn,10pt]{asme2ej}

\usepackage{graphicx} %% for loading jpg figures
\usepackage{hyperref}   % to set up hyperlinks
\hypersetup{
	colorlinks=true,
	linkcolor=blue,
	citecolor=blue,
	urlcolor=blue,
}
\usepackage[square,numbers]{natbib}

\usepackage{amsmath,amssymb,amsfonts}
\usepackage{graphicx}
\usepackage{textcomp}
\usepackage{cases}
\usepackage{algorithm}
\usepackage{algpseudocode}
\usepackage{array}
\usepackage{multirow}
\usepackage{hhline}
\renewcommand{\arraystretch}{1.2}
\usepackage{booktabs}
\usepackage{float}

\title{Integration of Multi-Mode Preference into Home Energy Management System Using Deep Reinforcement Learning}

\author{Mohammed Sumayli\\
     Dept. of Electrical and Computer Engineering\\
    FAMU-FSU College of Engineering\\
    Florida State University\\
	Tallahassee, Florida 32310\\
    Email: mas21bd@fsu.edu
}

\author{Olugbenga Moses Anubi\thanks{Address all correspondence related to ASME style format and figures to this author.}\\
    Dept. of Electrical and Computer Engineering\\
    FAMU-FSU College of Engineering\\
    Florida State University\\
	Tallahassee, Florida 32310\\
        Email: oanubi@fsu.edu
}

\begin{document}

\maketitle    

%%%%%%%%%%%%%%%%%%%%%%%%%%%%%%%%%%%%%%%%%%%%%%%%%%%%%%%%%%%%%%%%%%%%%%
\begin{abstract}
{\it Home Energy Management Systems (HEMS) have
emerged as a pivotal tool in the smart home ecosystem,
aiming to enhance energy efficiency, reduce costs, and im-
prove user comfort. By enabling intelligent control and
optimization of household energy consumption, HEMS
plays a significant role in bridging the gap between con-
sumer needs and energy utility objectives. However, much
of the existing literature construes consumer comfort as
a mere deviation from the standard appliance settings.
Such deviations are typically incorporated into optimiza-
tion objectives via static weighting factors. These factors often
overlook the dynamic nature
of consumer behaviors and preferences. Addressing this
oversight, our paper introduces a multi-mode Deep Re-
inforcement Learning-based HEMS (DRL-HEMS) frame-
work, meticulously designed to optimize based on dy-
namic, consumer-defined preferences. Our primary goal
is to augment consumer involvement in Demand Response
(DR) programs by embedding dynamic multi-mode pref-
erences tailored to individual appliances. In this study,
we leverage a model-free, single-agent DRL algorithm to
deliver a HEMS framework that is not only dynamic but
also user-friendly. To validate its efficacy, we employed
real-world data at 15-minute intervals, including metrics
such as electricity price, ambient temperature, and ap-
pliances’ power consumption. Our results show that the
model performs exceptionally well in optimizing energy
consumption within different preference modes. Furthermore, when compared to traditional algorithms based on
Mixed-Integer Linear Programming (MILP), our model
achieves nearly optimal performance while outperform-
ing in computational efficiency.  
}
\end{abstract}

%%%%%%%%%%%%%%%%%%%%%%%%%%%%%%%%%%%%%%%%%%%%%%%%%%%%%%%%%%%%%%%%%%%%%%

%%%%%%%%%%%%%%%%%%%%%%%%%%%%%%%%%%%%%%%%%%%%%%%%%%%%%%%%%%%%%%%%%%%%%%
\section{Introduction}

Home Energy Management Systems (HEMS) have played a crucial role in optimizing and influencing household energy consumption through the implementation of Demand Response (DR) mechanisms and integration with smart grid technologies \cite{taik2021demand}. DR is an energy management strategy that enables utility companies to actively modify consumer electricity usage patterns during peak demand periods. This strategy is often achieved through financial incentives or pricing signals provided to consumers to reduce or shift their electricity consumption during times of high demand or limited supply \cite{de2020demand}. One of the key drivers behind the development of HEMS is the implementation of smart grid \cite{chanpiwat2020using}. Smart grid incorporate bidirectional communication between electricity consumers and providers, allowing for real-time information exchange. This communication helps in effectively managing electricity demand and supply, leading to a reduction in renewable energy curtailment, and improving the overall reliability and efficiency of the grid \cite{piette2009design}. The integration of DR and smart grid technologies into HEMS allows for more sophisticated and dynamic control of energy consumption in residential buildings, where the main objective is to minimize electricity expenses for users within their desired and comfortable settings \cite{singer2019energyplus}.

In the past few decades, several research projects have been conducted to improve HEMS using conventional approaches to decrease peak demand. However, it is worth noting that this reduction may result in user discomfort, hence discouraging residents from actively engaging in DR programs \cite{han2011more, kundu2021stochastic}. The literature review conducted in \cite{han2023home} examines previous studies and applications of HEMSs that have used various optimization techniques. Many studies have developed price-based HEMSs that take consumer comfort into consideration. These studies employ distinct optimization strategies: classical methods, which encompass mixed integer linear programming \cite{duman2021home} and mixed integer nonlinear programming \cite{samadi2020home}; dynamic programming as highlighted in \cite{jeddi2019differential}; meta-heuristic methods, which comprise genetic algorithms \cite{sharifi2019energy}, particle swarm optimization \cite{carrasqueira2017bi} and butterfly optimization \cite{wang2021multi}; and finally, model predictive control as in \cite{jin2017foresee}. However, these techniques face some limitations and challenges such as the accuracy and complexity of building thermal dynamic models,  uncertainties, operational constraints, computational time, and dynamic environments \cite{yu2021review}.  

Reinforcement Learning (RL) is a decision-making paradigm where agents learn by interacting with their environment to maximize cumulative rewards \cite{kaelbling1996reinforcement}. One of the distinct advantages of RL is its ability to learn from trial and error without explicit supervision, enabling it to operate in environments where labeled data is scarce or nonexistent. Additionally, model-free RL methods, such as Q-learning and policy gradient methods, bypass the need for an explicit model of the environment's dynamics, allowing agents to generalize across different scenarios without being restricted to a specific model \cite{mnih2015human}. However, a key disadvantage of RL, especially in high-dimensional and complex environments, is its often substantial sample inefficiency. In such settings, agents may require vast amounts of interactions to learn even simple behaviors \cite{dulac2019challenges}. Deep Reinforcement Learning (DRL) has emerged as a powerful confluence of RL techniques and the representational power of Deep Learning (DL) models \cite{mnih2015human}. This union has enabled the automation of decision-making processes in complex, high-dimensional environments that were previously challenging for traditional RL methods. DRL's success has been particularly pronounced in domains such as game playing, with notable accomplishments in surpassing human performance \cite{silver2016mastering}. Beyond games, DRL has found promising applications in diverse fields like robotics and autonomous vehicles \cite{isele2018navigating}, showcasing its versatility and potential for real-world impact. 

 Many studies have proposed DRL algorithms to optimize energy consumption while satisfying user comfort regarding different home appliances and assets such as heating ventilation and air conditioning (HVAC) \cite{wei2017deep} HVAC and energy storage system (ESS) \cite{yu2019deep}. In \cite{lee2020energy}, a two-level DRL framework was proposed for scheduling home appliances such as air conditioner (AC) and washing machine (WM) at the first level based on consumer preferences and comfort and determining ESS and electric vehicle (EV) charging and discharging schedules at the second level using the optimal solution from the first level. The authors in \cite{amer2022drl} proposed a single-agent multi-objectives model that optimizes the energy consumption of home appliances and EV in the presence of ESS and PV generation from the consumer side. Some studies have applied multi-agent DRL where each agent controls one of the appliances within the same environment such as work presented in \cite{lee2019reinforcement}. However, the studies mentioned above, along with most related research, depict consumer discomfort as a deviation from the default settings of each home appliance and asset. This deviation is then incorporated into the objective function through specific weighting factors. These factors act as penalties, striking a balance between reducing electricity bills and ensuring consumer comfort. Notably, these weighting factors remain static and do not adapt to the evolving consumer behaviors and preferences over time. In \cite{nagy2018deep}, the authors reveal that the efficacy of model-free DRL agents declines when consumers modify the set points of the heat pump. This decline is attributed to the existing state-action pairs no longer aligning with the newly updated value functions within the agent's learning algorithm.

Consumers might express concerns about automated demand response systems, fearing discomfort from automated adjustments in heating or cooling and worrying about privacy and data security concerns. Additionally, feeling a loss of control over their appliances and energy consumption poses a barrier \cite{parrish2020systematic}. One widely used technique for evaluating thermal comfort is the Predicted Mean Vote (PMV) model, initially suggested by \cite{fanger1970thermal}, which forecasts the average thermal sensation ratings from a group of individuals using a seven-point scale \cite{standard1992thermal}. Nevertheless, this approach tends to be more effective in regulated office settings rather than in residential contexts, where the diverse temperature preferences of inhabitants and different assessment criteria for temperature come into play \cite{lissa2021deep}. Given the challenges in measuring PMV parameters and its oversight of subjective user feedback, its accuracy may be compromised. Authors in \cite{lissa2021deep} employ statistical analysis of indoor temperatures to determine dynamic temperature setpoints, relying on historical satisfaction and homeowner-approved adjustments to define demand response thresholds. Yet, this method's dependence on past data and predetermined adjustments may overlook the variability in daily occupancy and individual comfort preferences, potentially restricting the real-time adaptability and efficacy of demand response initiatives. Researchers emphasize the importance of setting clear thresholds before implementing a strategy \cite{sweetnam2019domestic,aghniaey2018impact}. 
 Hence, designing automated systems that consider these aspects, possibly through adjustable settings for user-defined comfort levels, is crucial \cite{parrish2020systematic}.

In this paper, we introduce a multi-mode DRL-HEMS framework, tailored to optimize based on consumer-defined preferences. The primary objective of this model is to bolster consumer engagement in DR programs by seamlessly integrating multi-mode preferences for individual home appliances into HEMS. Significantly, these preferences are dynamic, allowing adjustments at any given time to align with consumer comfort and convenience. Distinctively, to the best of our knowledge, this represents the pioneering effort in integrating multi-mode preferences within HEMS using a model-free, single-agent DRL approach, setting it apart from existing literature. Specifically, the contributions of this paper are enumerated as follows:
\begin{enumerate}
    \item A dynamic DRL-HEMS model, adept at adapting to evolving user 
          preferences for each appliance and load in real-time.
    \item An intuitive and flexible framework enabling users to transition 
          between overarching modes for all appliances and loads, or select distinct modes for individual appliances, ensuring ease of adaptability.
    \item A comprehensive evaluation leveraging authentic datasets, 
          encompassing electricity prices, ambient temperature, and power consumption metrics from sources like the Pecan Street database.
    \item Comparative analysis with the MILP method \textcolor{black}{shows that while our model closely approaches the optimal energy cost achieved by MILP, it significantly outperforms in computational efficiency.}
\end{enumerate}

The structure of this paper unfolds as follows: Section II delineates the architecture of the Home Energy Management System (HEMS), along with the mathematical modeling of household appliances and EV. Section III elaborates on the proposed Deep Reinforcement Learning-based HEMS (DRL-HEMS) methodology. Simulation outcomes are showcased in Section IV. The paper concludes with Section V, where we summarize our findings and outline avenues for future research.

\section{Overview of RL}

\subsection{Preliminary}
RL is a decision-making process that involves an agent interacting with an environment in a series of episodes making sequential decisions to maximize a carefully designed reward signal. Each episode unfolds over a series of timesteps. At each timestep, the agent perceives the current state of the environment and acts accordingly. Through these actions, the agent earns rewards.
This process is often modeled as a Markov Decision Process (MDP) which provides a mathematical framework for decision-making in an environment that is influenced by random events and where the immediate future states and action consequences depend only on the current state of the system.
This process persists until a terminal state is reached. Once trained using an offline dataset, the model can operate in a real-world setting. Should there be any changes in the environment, the agent adapts its actions, interacting with the altered environment to maximize rewards and maintain optimal performance.
By formulating the HEMS as an MDP, it becomes possible to effectively manage and control the energy usage of residential households in real time.

\subsection{Deep Q Networks}

Deep Q Networks (DQN) represents a significant advancement in the field of RL, blending traditional Q-learning with deep neural networks \cite{mnih2015human}. The foundation of Q-learning and, by extension, DQN, is the Bellman equation, which is formulated as:

\begin{equation}
    Q^{*}(s, a) =  r + \gamma 
    \max_{a^{\prime}} Q^{*}\left(s^{\prime}, a^{\prime}\right)
\end{equation}

Here, $Q^{*}(s, a)$ represents the optimal action-value function, which gives the expected utility of taking action $a$ in the current state $s$ and then following the optimal policy. The immediate reward after taking action $a$ in state $s$ is denoted by $r$, while $\gamma$ is the discount factor, highlighting the distinction between the values of immediate and future rewards. The term $\max_{a^{\prime}} Q^{*}\left(s^{\prime}, a^{\prime}\right)$ represents the maximum expected utility achievable by taking some action $a^{\prime}$ in the new state $s^{\prime}$, under the optimal policy.

The methodology underlying DQN is underpinned by two primary mechanisms: experience replay and target network. The former involves storing transition tuples ($s, a, r, s^{\prime}$) in a replay buffer and then randomly sampling from this buffer during training. This random sampling helps to de-correlate consecutive samples, enabling more stable and efficient learning. On the other hand, the target network strategy introduces a secondary Q-network with frozen parameters to compute the target Q-values during updates. This network's parameters are updated less frequently than the primary Q-network, which aids in mitigating oscillations and instabilities in learning. Formally, the Q-value update equation in DQN can be represented as:

\begin{multline}
Q^{k+1}(s, a; \theta) = Q^k(s, a; \theta) \\ + \alpha \left[r + \gamma \max_{a^{\prime}} Q^k(s^{\prime}, a^{\prime}; \theta^-) - Q^k(s_, a; \theta)\right]
\end{multline}

where $\alpha$ is the learning rate, determining the weight given to the new information. $\theta$ and $\theta^{-}$ are the weights of the primary and target networks, respectively. However, standard DQN tends to overestimate action values, leading to suboptimal policy decisions.
This overestimation occurs because the same network ($\max_{a'} Q^t(s', a'; \theta^-)$) is used for both selecting and evaluating an action, leading to a positive bias. This bias can propagate through updates, resulting in a significant overestimation of Q-values.

To counter this, the Double DQN was introduced by \cite{van2016deep}. The key idea behind Double DQN is to decouple the action selection from the action evaluation, where two sets of weights are used: one for selecting the best action in the next state ($\theta$) and another set of weights ($\theta^-$) for evaluating the value of taking that action. Therefore, the iteration update of the Q-value becomes as follows:

\begin{align}\nonumber
Q^{k+1}(s, a; \theta) &= Q^k(s, a; \theta) + \alpha \Bigg[r \\\nonumber
                      &\quad + \gamma Q^k\Big(s^{\prime}, \underset{a^{\prime}}{\text{argmax}} Q^k\big(s^{\prime}, a^{\prime}; \theta\big); \theta^{-}\Big) \\
                      &\quad -  Q^k(s, a; \theta)\Bigg]
\end{align}

Furthermore, the standard DQN algorithm exhibits certain limitations in particular environments or scenarios where the value of being in a certain state does not depend heavily on the action chosen. This is especially true in states where the choice of action has minimal impact on what happens next. In such scenarios, the standard DQN, which combines the estimation of state values and action advantages into a single stream, may struggle to differentiate between the value of the state itself and the value of actions within that state, leading to inefficiencies in learning and policy evaluation. To deal with this issue, dueling network architecture was proposed by \cite{wang2016dueling}. 
Dueling DQN introduces a nuanced structural modification to the neural network model used in DQN. It separates the network into two streams: one for estimating the state value function 
$V(s)$, which gauges the value of being in a particular state, and another for the advantage function 
$A(a)$, which assesses the additional benefit of each action in that state. The final Q-value is then computed by combining these streams, adjusting the advantage of each action by subtracting the mean advantage of all actions in that state. 

\begin{equation}
\begin{aligned}
& Q(s, a ; \theta, \alpha, \beta)=V(s ; \theta, \beta)+ \\
& \qquad\left(A(s, a ; \theta, \alpha)-\frac{1}{|\mathcal{A}|} \sum_{a^{\prime}} A\left(s, a^{\prime} ; \theta, \alpha\right)\right)
\end{aligned}
\end{equation}

where $\beta$ and $\alpha$ are the parameters of the value and advantage streams of the network, respectively, and 
${|\mathcal{A}|}$ is the number of possible actions. 

Table \ref{tab:DQN_comparison} provides a comparative analysis of DQN, Double DQN, and Dueling Double DQN, highlighting their strengths and limitations

\begin{table*}[h]
    \centering
    \caption{Comparison of DQN, Double DQN, and Dueling Double DQN}
    \renewcommand{\arraystretch}{1.3}
    \begin{tabular}{m{2.8cm} m{3.5cm} m{4cm} m{4cm}}
        \toprule
        \textbf{Feature} & \textbf{DQN} & \textbf{Double DQN} & \textbf{Dueling Double DQN} \\
        \midrule
        \textbf{Overestimation Bias} & High  & Reduced (decoupled action selection \& evaluation) &  Reduced (inherits Double DQN improvements) \\
        \textbf{Architecture} & Single Q-network & Two Q-networks (online \& target) & Two Q-networks with value/advantage separation \\
        \textbf{Learning Stability} & Moderate & Improved due to bias reduction & Most stable due to combined bias reduction and value learning \\
        \textbf{Decision-Making Performance} & Prone to overestimation & More stable decisions, avoiding unnecessary actions & Best trade-off between stability and efficiency \\
        \textbf{Computational Complexity} & Moderate & Slightly higher due to two networks & Highest due to combined Double \& Dueling structures \\
        \bottomrule
    \end{tabular}
    \label{tab:DQN_comparison}
\end{table*}

%%%%%%%%%%%%%%%%%%%%%%%%%%%%%%%%%%%%%%%%%%%%%%%%%%%%%%%%%%%%%%%%%%%%%%
\section{HEMS Framework and Modeling of Home Appliances}

%%%%%%%%%%%%%%%%%%%%%%%%%%%%%%%%%%%%%%%%%%%%%%%%%%%%%%%%%%%%%%%%%%%%%%
\subsection{HEMS Framework}

\begin{figure}
    \centering
    \includegraphics[width=1\linewidth]{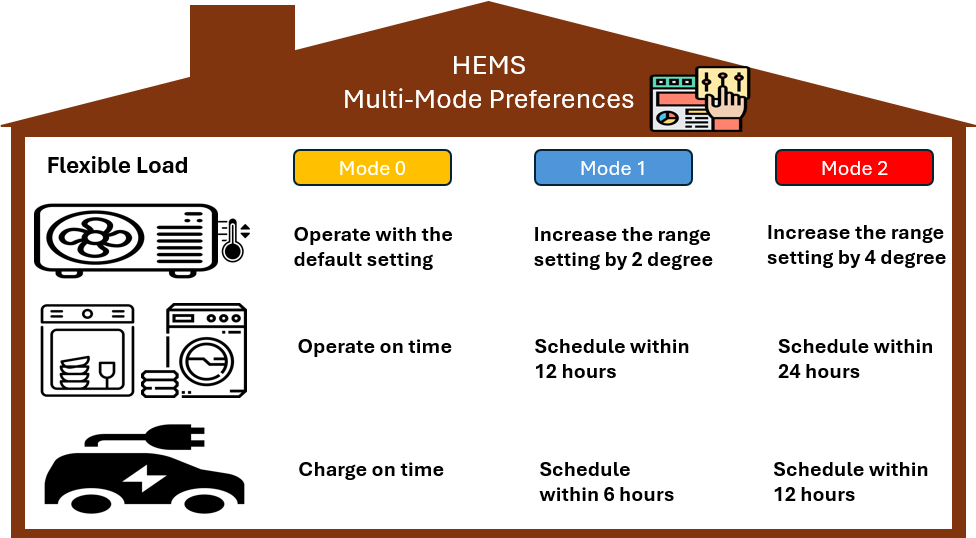}
    \caption{Framework of multi-mode HEMS. Icons sourced from Flaticon (https://www.flaticon.com).}
    \label{fig:HEMS}
\end{figure}

The HEMS framework presented in Fig \ref{fig:HEMS} is designed to streamline and optimize energy consumption across a suite of household appliances and EV. Central to the HEMS is the introduction of three distinct operational modes, each representing varying levels of flexibility for the DR program. Mode zero serves as the baseline, functioning as the default setting for all appliances, with no active DR interventions. Mode one introduces a moderate level of flexibility: the temperature set points for regulatable appliances such as HVAC can be adjusted by a margin of 1 degree Celsius either way \textcolor{black}{(a total range of 2 degrees Celsius, meaning the upper limit is 1 degree above the setpoint and the lower limit is 1 degree below the setpoint)}, while the time-shiftable appliances are provided a 12-hour operational window. Concurrently, EV under this mode are allocated a 6-hour charging timeframe. Mode two escalates the adaptability further, signaling heightened consumer engagement with DR programs aimed at more financial savings. Here, the HVAC's temperature set point range broadens by two degrees \textcolor{black}{(a total range of 4 degrees Celsius, meaning the upper limit is 2 degrees above the setpoint and the lower limit is 2 degrees below the setpoint)}, the time-shiftable appliances expand their task completion window to 24 hours, and the EV receives a 12-hour charging interval. Through these modes, the HEMS aspires to harmoniously balance energy efficiency with consumer preferences and comfort.

%%%%%%%%%%%%%%%%%%%%%%%%%%%%%%%%%%%%%%%%%%%%%%%%%%%%%%%%%%%%%%%%%%%%%%

\subsection{Modeling of Home Appliances and EV}

Consider a HEMS where time-shiftable appliances, controllable appliances, and EV are each integrated with a set of preference modes, denoted by \( m \in \{\text{mode 0}, \text{mode 1}, \text{mode 2}\} \). These appliances are classified based on their operational characteristics and consumption patterns. Each category is defined by its unique operational state. Given that the real-time electricity price is intricately linked to the day-ahead market price \cite{huisman2007hourly}, the average day-ahead price has been incorporated into the operational states of all appliances, reflective of their current preference mode.

\subsubsection{Time-Shiftable Appliances.}

These appliances offer flexibility in adjusting their operation time. However, once they commence their operational cycle, they cannot be interrupted. Consider an assortment of \( N \) time-shiftable appliances where each appliance is indexed by \( n \in \{1, 2, \ldots, N\} \), the state of the shiftable appliance ${SA}_\text{n}(t)$ is defined as:

\begin{equation}
{SA}_\text{n}(t) = \left[u_\text{n}(t),~ w_\text{n}(t),~ x_\text{n,m}(t),~ z_\text{n,m}\right]
\label{eq1}
\end{equation}

where $u_{\text {n}}(t)$ $\in\{0,1\}$ indicates whether or not the device $n$ has been activated by the user, taking the value 1 if it is. The appliance's performance is assessed by the value of $w_{\text {n}}(t) \in\mathbb{R}$. $x_{\text {n,m}}(t) \in\mathbb{R}$ is the remaining time of the scheduled window interval to start the appliance, and $z_{\text {n,m}} \in\mathbb{R}$ is the average prediction price of the scheduled window interval. For instance, when a user activates a shiftable appliance, referred to as appliance $n$, by selecting either mode 1 or mode 2, we calculate the average predicted price for the scheduled window interval using the day-ahead energy price. Specifically, this average is based on the forthcoming 12 or 24 hours of the day, aligning with the duration specified by the chosen mode.
Consider a shiftable appliance requiring a duration of $d$ time slots to finish a cycle. The operation of the \( n \)-th time-shiftable appliance is characterized by the variables \( t_{a,n} \), \( t_{\text{start}, n} \), and \( t_{b,n,m} \), where \( t_{a,n} \) is the activation time, \( t_{\text{start}, n} \) is the actual start time, and \( t_{b,n,m} \) is the end time of the scheduled window interval for mode \( m \). These variables are subject to the box constraint \( t_{a,n} \leq t_{\text{start}, n} \leq t_{b,n,m}-d_n \). Given this constraint, the state  of the appliance can be described as:

\begin{equation}
\begin{aligned}
&\left[u_n(t),~ w_n(t),~ x_{n,m}(t),~ z_{n,m}\right]\\ & = \begin{cases}\left[1,~ \textcolor{black}{w_n(t)},~ \textcolor{black}{x_{n,m}(t)},~ z_{n, m}\right],&\scalebox{0.9}{$ t \in\left[t_{a, n}, \textcolor{black}{t_{start, n}+d_{n}}\right]$} \\
[0,~ 0,~ 0,~ 0],& \text { otherwise }\end{cases}
\label{eq2}
\end{aligned}
\end{equation}

\textcolor{black}{The state transition for the operation progress $w_{\text {n}}(t)$ is given by \eqref{w}, where  $k_{n}(t)$ is a binary control variable indicating if the appliance is operating at time $t$. As $k_{n}(t)$  accumulates over the duration $d_{n}$, $w_{\text {n}}(t)$ reflects the normalized progress of the appliance's operation. Initially, $w_{\text {n}}(t) = 0$ and it increases as the appliance operates, reaching 1 when the cycle is complete.}

\textcolor{black}{
\begin{equation}
    {w_n(t)} = \frac{\sum_{i=t-d_{n}+1}^{t} k_n(i)}{d_n}
    \label{w}
\end{equation}}

\textcolor{black}{ The control variable $k_{n}(t)$} dictates the operational and time constraints specified for the appliance, as defined below:

\begin{subnumcases}{k_{n}(t)=}
1, \quad { if } ~ k_{n}(t-1)=1 ~ { and }   ~ 0<w_{n}(t)<1 
\label{eq4a}
\\
1, \quad { if } ~ \textcolor{black}{u_{n}(t) = 1 ~ { and } ~  x_{n,m}(t) = 0} \label{eq4b} \\
0, \quad { if } ~t ~ \notin\left[t_{a, n}, t_{b, n, m}\right]   \label{eq4c}\\
\textcolor{black}{{a}_\text{SA,n}(t), \quad {otherwise }} \label{eq4d}
\end{subnumcases}

Specifically, \eqref{eq4a} ensures that the appliance can run without interruption, \eqref{eq4b} ensures that it will start the task before the time limit, \eqref{eq4c} limits the action to 0 when the appliance is not in its active window, \textcolor{black}{and \eqref{eq4d} is to ensure that the backup controller ($k_{n}(t)$) align with the optimization controller represented by ${a}_\text{SA,n}(t)$.}

\textcolor{black}{The state transition of $x_{\text {n,m}}(t)$ described in \eqref{x}, starts at $t = t_{a,n}$ with an initial value $(t_{b,n,m} - t_{a,n}) -d_n$, representing the total window length minus the duration of the appliance's cycle. This value decreases by 1 at each time step, reflecting the countdown until the appliance completes its operation cycles.}

\textcolor{black}{
\begin{equation}
   x_{n,m}(t) = 
(t_{b,n,m} - t_{a,n}) - d_n - (t - t_{a,n})
\label{x}
\end{equation}}

\subsubsection{Controllable Appliances} For appliances of this kind, for example the HVAC, the power consumption is adjustable. The state of the HVAC is defined as follows:

\begin{equation}
{HVAC}(t) = [T_{\text {in}}(t) ,  ~  T_{\text {out}}(t) ,  ~  T_{\text {max,m}} ,  ~ T_{\text {min,m}}, ~ z_{\text {HVAC,m}}(t)]
\label{eq4}
\end{equation}

where $T_\text{in}\in\mathbb{R}$ the inside temperature, $T_{\text {out}}\in\mathbb{R}$ is the outside temperature, and $T_{\text {max,m}}$, $T_{\text {min,m}}$ are the upper and lower limits of the temperatures range of the thermostat settings based on the chosen preference mode $m$. For appliances of this category, the average predicted price, denoted as $z_{\text {HVAC,m}}(t)$, varies with time $t$ depending on the chosen mode $m$. Specifically, for mode 1, we calculate the average energy price projected for the upcoming two hours. In contrast, for mode 2, the calculation extends to cover the next four hours. The indoor temperature of the simplified equivalent thermal parameters (ETP) model  \cite{lu2012evaluation} given by:

\begin{multline} 
T_{\text{in}}(t+1)=T_{\text{out}}(t+1) \\ +Q R-\left(T_{\text{out}}(t+1)+Q R-T_{\text {in}}(t)\right) e^{-\Delta t / R C}
\label{eq5}
\end{multline}

where Q, R, and C $ \in \mathbf{R}$ are the equivalent heat rate, equivalent thermal resistance, and equivalent heat capacity respectively. This model follows the first law of thermodynamics, which states that the rate of energy change in a system equals the net heat transfer. Based on this principle, the thermal dynamics of the building can be expressed as:

\begin{equation}
C \frac{dT_{\text{in}}}{dt} = \frac{T_{\text{out}} - T_{\text{in}}}{R} - Q,
\end{equation}

where the left-hand side represents the rate of change of indoor temperature, while the right-hand side accounts for the heat exchange between the indoor and outdoor environment and the HVAC heat rate. Equation~\eqref{eq5} is a discrete-time representation of this relationship, where the term \( Q R \) captures the steady-state contribution of the HVAC system, and \( e^{-\Delta t / (R C)} \) accounts for transient thermal dynamics, reflecting the thermal inertia of the building.
\textcolor{black}{While the power consumption can be adjusted based on environmental conditions, our model uses a binary control variable $a_\text{HVAC}$ to turn the HVAC on or off at each time step (15 minutes), where the heat rate $Q(t)$, for cooling or heating, is derived at each time step by bringing the indoor temperature to the setpoint $T_{set}$, considering both the indoor and outdoor temperatures, while also respecting the physical constraints, such as the upper and lower bounds of the heat rate.
The heat rate $Q(t)$ can be calculated as follows}:

\begin{equation} Q(t) = \frac{T_{\text{set}} - T_{\text{out}}(t) + \left(T_{\text{out}}(t) - T_{\text{in}}(t-1)\right) e^{-\frac{\Delta t}{R \cdot C}}}{R \cdot (1 - e^{-\frac{\Delta t}{R \cdot C}})} \label{eqQ} \end{equation}

\textcolor{black}{ where}

\textcolor{black}{
\begin{equation}
    -Q_{\text{max}} \leq Q(t) \leq Q_{\text{max}}
    \label{Qmax}
\end{equation}}

Thus, the power consumption $p_\text{HVAC}(t)$ can be calculated as \cite{bojic2011simulation, pospivsil2018potential}:

\textcolor{black}{
\begin{equation}
    p_{\text{HVAC}}(t) = \frac{Q(t) \cdot a_{\text{HVAC}}(t)}{\text{CoP}}
    \label{PHVAC}
\end{equation}}

\textcolor{black}{where CoP is the coefficient of performance and $Q_{max}$ is the maximum limit of the heat rate.} For the simulations, we used a CoP value of 3.5 and $Q_{max}$
 =14kW, as in \cite{katipamula2006evaluation}.

\textcolor{black}{Therefore, the indoor temperature can be updated based on our model as follows:}

\textcolor{black}{
\begin{multline} 
T_{\text{in}}(t)=T_{\text{out}}(t)  +Q(t)~a_{\text{HVAC}}(t)~ R \\ -\left(T_{\text{out}}(t)+Q(t)~ a_{\text{HVAC}}(t)~ R-T_{\text {in}}(t-1)\right) e^{-\Delta t / R C}
\label{T_indoor}
\end{multline}}

\subsubsection{Electric Vehicle}
Controlling the charging duration of an EV is possible by satisfying certain constraints. Assuming the EV enters the system with minimum State of Charge of its battery ($SoC = SoC_{\text {min}}$) at time $t_{arr}$ and is designated to be fully charged ($SoC = SoC_{\text {max}}$) at time $t_{dep, m}$, we may define its operation state as follows:

\begin{equation}
{EV}(t) = 
\left[u_\text{EV} (t),~ SoC\text (t),~ x_{\text {EV,m}}(t), ~z_{\text {EV,m}}\right]
\label{eq6}
\end{equation}

where

\begin{equation}
SoC_{\text {min}} \le SoC\text (t) \le SoC_{\text {max}}
\label{eq7}
\end{equation}

\textcolor{black}{
\begin{equation}
    \text{SoC}(t) = \text{SoC}(t-1) + \frac{k_{\text{EV}}(t) \cdot p_{\text{EV}} \cdot \eta_{\text{EV}} \cdot \Delta t}{C_{\text{battery}}}
\end{equation}}

where ${p}_\text{EV}$ is the charging power rate in (kW), $\eta_{EV}$ is the charging efficiency, \textcolor{black}{and $C_\text{battery}$ is the capacity of the EV battery}. The
operational and time constraints specified for the EV is
defined as:

\begin{subnumcases}{k_\text{EV}(t)=}
1, & \text{if } $SoC(t) < SoC_\text{max}$ \notag \\
   & \text{and } $t = t_\text{dep,m} - d_{EV}(t)$ \label{eq14a} \\
0, & \text{if } $t \notin [t_\text{arr}, t_\text{dep,m}]$ \label{eq14b} \\
\textcolor{black}{a_{\text{EV}}(t)}, & \text{otherwise} \label{eq14c}
\end{subnumcases}

Specifically, \eqref{eq14a} guarantees the completion of the charging task within the designated time limit, \eqref{eq14b} restricts the action to zero in scenarios where the EV is not present in the system, \textcolor{black}{and \eqref{eq14c} optimizes the charging based on the agent's action.}

%%%%%%%%%%%%%%%%%%%%%%%%%%%%%%%%%%%%%%%%%%%%%%%%%%%%%%%%%%%%%%%%%%%%%%
\section{Methodology}

\subsection{MDP Formulation}
To model the decision-making process in our framework, we formulate the problem as MDP, where the agent interacts with the environment by selecting actions at each timestep to maximize a reward signal. We define the specific components that structure our problem, including the state representation, action space, and reward function.

\subsubsection{State}
The state represents the system’s conditions at any given timestep. In this study, the state vector at time 
$t$ is defined as:

\begin{equation}
{s}(t) = \left[{SA}_\text{1}(t),~ ...,~ {SA}_\text{N}(t),~ {HVAC}(t),~ {EV}(t),~ \rho(t) \right]
\end{equation}

where $\rho(t)$ is the energy price.

\subsubsection{Action}
The action set defines the possible control decisions for the agent. The action vector at time $t$ is given by: 

\begin{equation}
{a}(t) = \left[{a}_\text{SA,1}(t),~ ...,~ {a}_\text{SA,N}(t),~ {a}_\text{HVAC}(t),~ {a}_\text{EV}(t)\right]
\end{equation}

In this context $a_\text{SA}$, $a_\text{HVAC}$ and $a_\text{EV}$ represent binary actions executed by the agent, specifically aimed at optimizing the operational times of shiftable appliances, HVAC, and EV, respectively.

\subsubsection{Reward}
The reward function provides feedback to the agent, guiding it toward optimal decision-making. The total reward at time $t$ is defined as:

\begin{equation}
r(t) = \sum_{n=1}^{N} {r}_\text{SA,n}(t) + {r}_\text{HVAC}(t) + {r}_\text{EV}(t) ~,
\end{equation}

where

\begin{multline}
 {r}_\text{SA,n}(t) = ({z}_\text{n,m} - \rho(t)) {p}_\text{SA,n}(t)~ \textcolor{black}{{k}_\text{n}(t)}  \\+ |{a}_\text{SA,n}(t) - {k}_\text{n}(t) | ~\zeta_\text{SA}
\end{multline}

\begin{equation}
\begin{aligned}
r_\text{HVAC}(t) =
\begin{cases}
(z_\text{HVAC,m}(t) - \rho(t))p_\text{HVAC}(t), \\
\qquad \text{if } T_\text{in}(t) \in [T_\text{min,m}, T_\text{max,m}], \\
(T_\text{min,m} - T_\text{in}(t))\zeta_\text{CA}, \\
\qquad \text{if } T_\text{in}(t) < T_\text{min,m}, \\
(T_\text{in}(t) - T_\text{max,m})\zeta_\text{CA}, \\
\qquad \text{if } T_\text{in}(t) > T_\text{max,m}.
\end{cases}
\end{aligned}
\end{equation}

\begin{multline} 
 {r}_\text{EV}(t) = ({z}_\text{EV,m} - \rho(t)) {p}_\text{EV}(t)~ \textcolor{black}{{k}_\text{EV}(t)} \\ + |{a}_\text{EV}(t) - {k}_\text{EV}(t) |~ \zeta_\text{EV}
\end{multline}

Here, $\zeta_\text{SA}$ = -0.1, $\zeta_\text{CA}$ = -5 and $\zeta_\text{EV}$ = -0.1 represent respective penalty factors. The purpose of these penalty factors is to incentivize the agent to adhere to operational and comfort constraints specified by the user, according to the selected preference mode. These values were determined through iterative tuning to balance cost minimization and constraint satisfaction. Lower penalties for SA and EV allowed flexible scheduling since their constraints are enforced by backup mechanisms, while a higher penalty for HVAC ensured strict thermal comfort adherence due to the lack of such a mechanism. ${p}_\text{SA,n}(t)$ is the power consumption of the shiftable appliance $n$. The aim of this reward function is to operate the appliances and charge the EV at the lower price of their scheduled time interval.

%%%%%%%%%%%%%%%%%%%%%%%%%%%%%%%%%%%%%%%%%%%%%%%%%%%%%%%%%%%%%%%%%%%%%%
\subsection{DRL-HEMS}

In our study, we implemented the Dueling Double DQN algorithm, which effectively combines the strengths of Double DQN and Dueling DQN to optimize decision-making in complex environments. This integration is particularly significant in scenarios that require accurate action-value estimations, such as HEMS.

\begin{algorithm}
\label{alg}
\caption{ Dueling Double DQN}
\begin{algorithmic}[1] % The number indicates the line numbering frequency
\State Initialize replay memory $D$ 
\State Initialize main network $Q$  with random weights $\theta$
\State Initialize target network $Q$ with weights $\theta^- = \theta$

\For{episode = 1 to E}
    \State Initialize state $s$
    \While{$s$ is not terminal}
        \State $a = \begin{cases}\text{random action} , & \text{with } \epsilon \text{ probability} \\
        \underset{a}{\text{argmax}} Q(s, a; \theta), & \text{ otherwise }\end{cases}$
        
        \State Execute action $a$ in environment
        \State Observe $r$ and $s'$
        \State Store transition $(s, a, r, s')$ in $D$
        \State Sample random minibatch of transitions 
        \For{each transition $(s, a, r, s')$ in minibatch}
            \If{$s'$ is terminal}
                \State $y = r$
            \Else
                
                \State $\scalebox{0.95}{$y = r + \gamma Q\Big(s^{\prime}, 
                \underset{a^{\prime}}{\text{argmax}} Q\big(s^{\prime}, a^{\prime}; \theta\big); \theta^{-}\Big)$}$

            \EndIf
            \State Gradient descent step on
                $(y - Q(s, a; \theta))^2$ 
                \par \hspace*{2.3em} $w.r.t$ the network parameter $\theta$
        \EndFor
        \State Update $s == s'$
        \State After C steps,  $\theta^{-} == \theta$
    \EndWhile
\EndFor
\end{algorithmic}
\end{algorithm}

\begin{figure*}[h]
    \centering
    \includegraphics[width=1\linewidth]{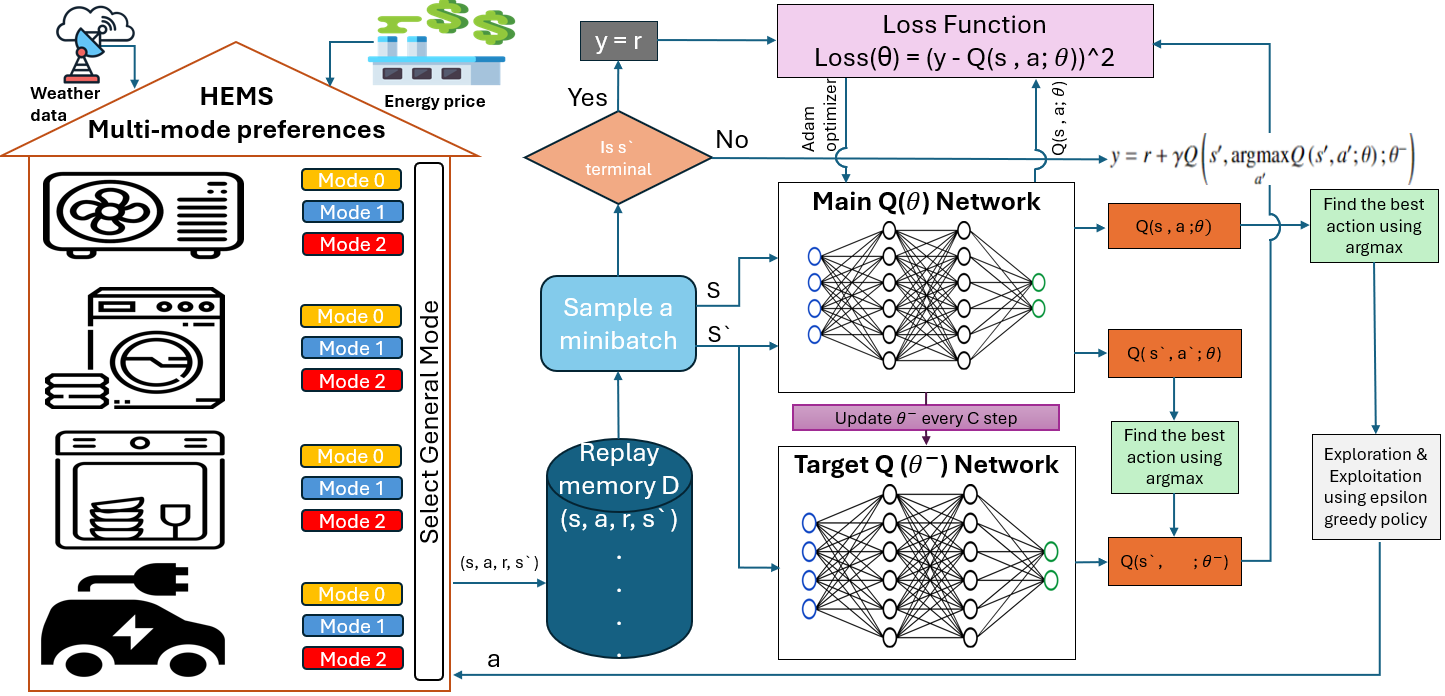}
    \caption{Structural diagram of the Dueling Double DQN algorithm. Icons sourced from Flaticon (https://www.flaticon.com).}
    \label{fig:HEMS-DDDQN}
\end{figure*}

The Dueling Double DQN algorithm, presented in Algorithm 1, begins with the establishment of essential foundational structures: initializing a replay memory $D$ to retain experience tuples, while concurrently setting up both the main Q-network and the target Q-network, with commencing weights $\theta$ and $\theta'$ respectively (steps 1-3). Over a predefined number of training episodes (steps 4-23), the algorithm starts each episode by defining an initial state 
$s$ (step 5). Within the episode loop, until the state is terminal (steps 6-22), the algorithm makes its action selection (step 7) based on an $\epsilon$-greedy policy: it opts for a random action with a probability of $\epsilon$ or selects the action that maximizes the Q-value of the main network. Then, the chosen action is executed in the environment and results in an observed reward and a subsequent state $s'$, which together with the prior state and action, are cataloged as a transition in $D$ (steps 8-10). To facilitate nuanced learning from diverse experiences, a random subset of stored transitions is sampled (step 11).  For each sampled transition (steps 12-19), a target Q-value $y$ is determined: it's set as the observed reward if $s'$ is terminal (steps 13-14); otherwise, it incorporates the Q-value associated with the optimal action from the main network into the target network's assessment (steps 16). Through gradient descent, the online network updates its weights by minimizing the differential between the estimated and target Q-values (step 18). This iterative learning process advances by updating the state (Step 20), and periodically, for stability, the target network's weights are synchronized to mirror the main network after a specified number of steps (step 21). \textcolor{black}{Fig \ref{fig:HEMS-DDDQN} shows the structural diagram of the algorithm.} The proposed model is implemented in Python, utilizing key libraries such as TensorFlow for deep learning, Gym for environment modeling, NumPy and Pandas for numerical computations, and Matplotlib for result visualization.

%%%%%%%%%%%%%%%%%%%%%%%%%%%%%%%%%%%%%%%%%%%%%%%%%%%%%%%%%%%%%%%%%%%%%%
\section{Simulation and Results}

In this work, appliances that contribute significantly to energy consumption such as washing machine (WM), dishwasher (DW), HVAC, and EV are considered to determine their DR and improve their operations over time to save energy costs. The power and time of consumption data for DW, WM, and EV have been taken from the \href{https://www.pecanstreet.org/dataport/}{Pecan Street database} for one year, \textcolor{black}{ensuring real-world representation of operational timing and power consumption uncertainties, including the EV's arrival time}. The EV has a charging power rate of 3.4 $kW$, while its battery has a rating of 17 $kWh$ with $SoC_{min}$ and $SoC_{max}$ of 20\% and 90\%, respectively. Assuming the EV arrives at home at the minimum level ($SoC = SoC_{\text {min}}$), the EV requires a total of 3.5 hours to complete its charging process, operating at its charging rate. \textcolor{black}{The arrival time marks the start of the charging process, while the departure time depends on the selected mode, a variable the consumer sets. During the scheduling window of mode 1 (6 hours) and mode 2 (12 hours), the EV can be charged at the rated power at any time}.
The HVAC system has been simulated using the ETP model as discussed in section II with typical values for the key input parameters obtained from \cite{katipamula2006evaluation} for a residential house with an area of 2,000 square feet. \textcolor{black}{Given that the setpoint temperature $T_{set}$ for the HVAC system is fixed at 23°C, with mode 0 using a default deadband of 0.5°C, mode 1 extends the deadband to 2°C, resulting in $T_{\text{min,m}}$ = 22°C and $T_{\text{max,m}}$ = 24°C. Mode 2 further expands the deadband to 4°C, setting $T_{\text{min,m}}$ = 21°C and $T_{\text{max,m}}$ = 25°C.} These temperature ranges align with established comfort models, such as ASHRAE Standard 55 \cite{ASHRAE55}, which considers indoor temperatures between 20°C and 26°C acceptable for residential settings under adaptive comfort conditions. The flexibility of these settings allows users to define their comfort preferences, and the model adapts accordingly to optimize energy efficiency while maintaining comfort. Furthermore, for the HVAC system, the average prediction price states for modes 1 and 2 are computed based on two-hour and four-hour intervals of the day-ahead energy price, respectively. 
In this study, our action set comprises four binary actions, signifying whether the appliance is either "on" or "off."

Simulation has been conducted with 15-minute timestep intervals using real-world electricity price data from \href{https://www.nyiso.com/energy-market-operational-data}{NYISO} for the year 2022. The one-year dataset was divided into two sets, with data from January to October used for training while the remaining two months' data was used for testing. The length of the training episode is two days (48 hours) which is equivalent to 192 timesteps where these two days are chosen randomly from the training dataset for each episode. Table \ref{tab:t1} presents the hyper-parameters that were used to train the model.
Preference modes were selected randomly for each appliance at the beginning of each episode.

\begin{table}[t]
  \centering
  \caption{Hyper-parameters of the algorithm}
  \footnotesize % Reduce font size to fit table in one-column format
  \setlength{\tabcolsep}{3pt} % Reduce column spacing

  \begin{tabular}{m{1.3cm} m{0.8cm} m{5cm}} 
\toprule[1.5pt] \textbf{Description} & \textbf{Value} & \textbf{Justification} \\
\midrule[1.5pt] 
Number of episodes & 1500 & Ensures sufficient training for stable convergence. \\
Timesteps per episode & 192 & Matches a 48-hour scheduling horizon with 15-minute resolution. \\
Learning rate & 0.001 & Balances learning speed and stability, preventing divergence. \\
Discount factor & 0.99 & Prioritizes long-term rewards for optimal cost savings. \\
Epsilon (Max) & 1 & Enables full exploration at the start of training. \\
Epsilon (Decay) & 0.005 & Smoothly transitions from exploration to exploitation. \\
Epsilon (Min) & 0.01 & Retains minimal exploration to avoid local optima. \\
\bottomrule[1.5pt]
\end{tabular}

  \label{tab:t1}
\end{table}

\begin{figure}
    \centering
    \includegraphics[width=1\linewidth]{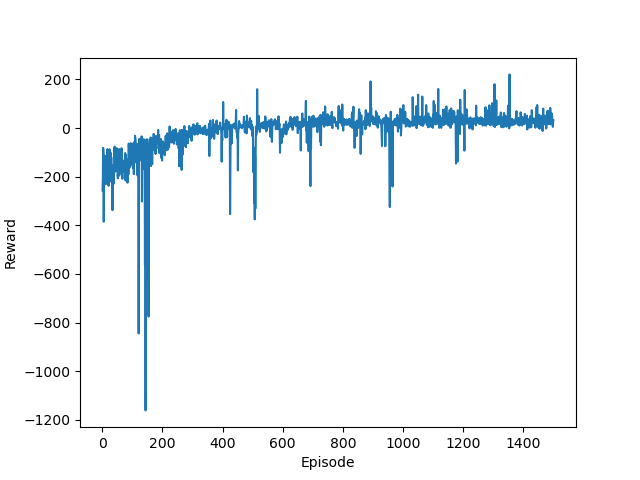}
    \caption{Rewards over the number of training episodes }
    \label{fig:reward}
\end{figure}

Throughout training, the Dueling Double DQN agent progressively acquires the ability to adapt to the environment and trigger bigger rewards. The training is shown in Fig \ref{fig:reward}. In the beginning, the agent takes numerous actions at random. As training progresses, it gradually learns to select actions that achieve the optimization objective. As seen in the figure, the rewards increase over time and converge after 1000 episodes. There are still variations in obtained rewards after convergence. This is due to the fluctuation of real-time electricity prices.

\begin{figure}
    \centering
    \includegraphics[width=1\linewidth]{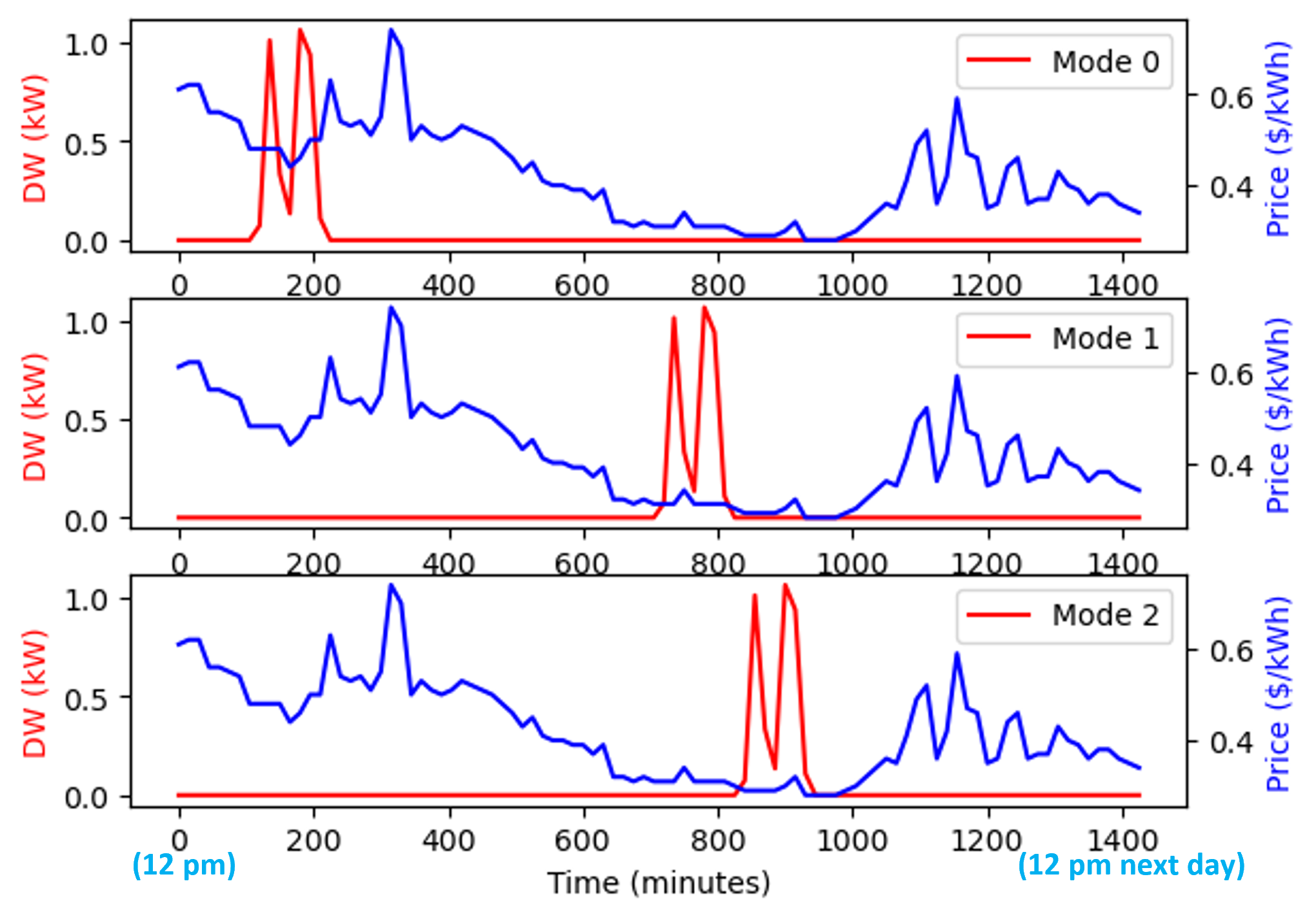}
    \caption{Scheduling of dishwasher with different preferences}
    \label{fig:DW}
\end{figure}

\begin{figure}
    \centering
    \includegraphics[width=1\linewidth]{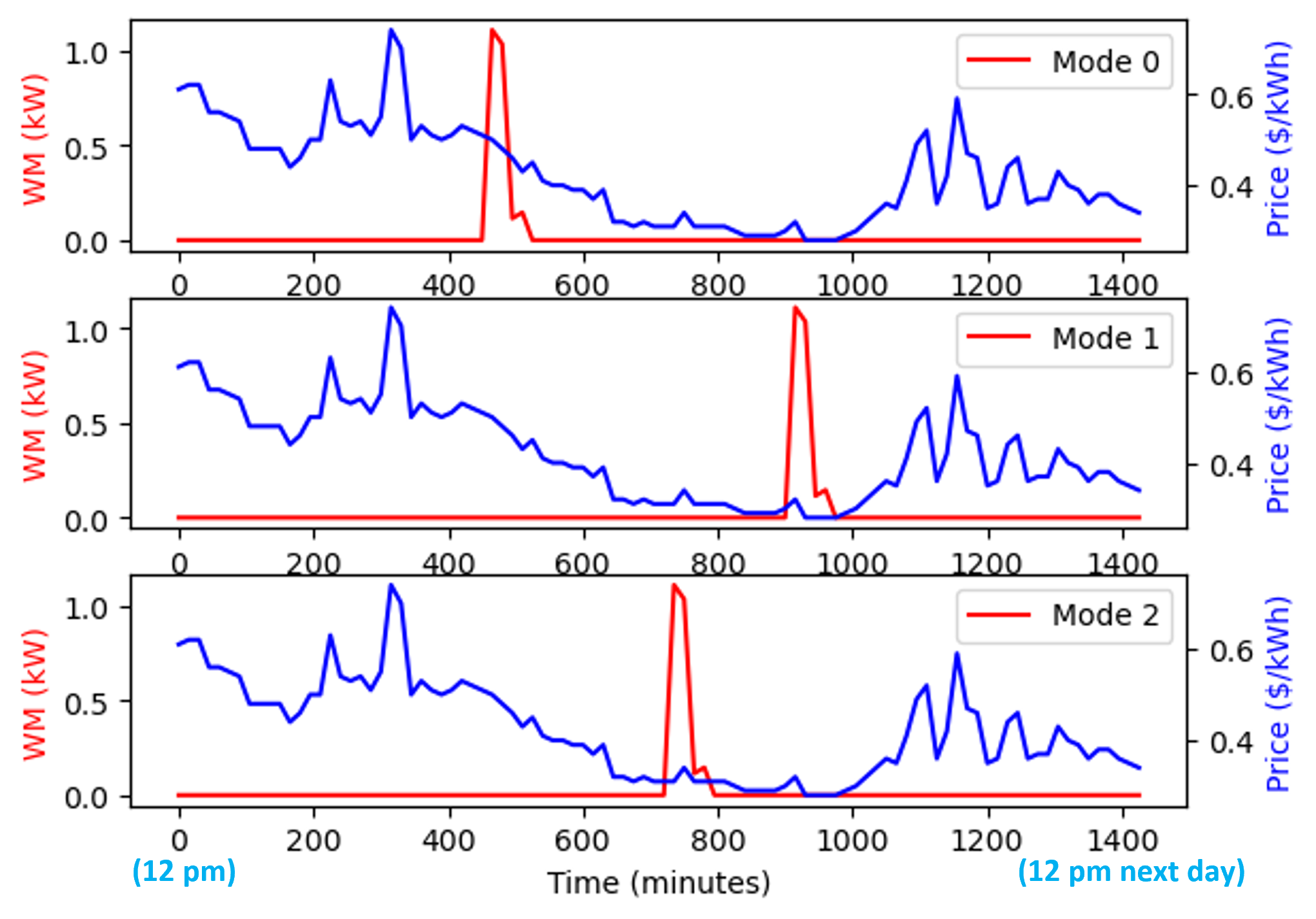}
    \caption{Scheduling of washing machine  with different preferences}
    \label{fig:WM}
\end{figure}

\begin{figure}
    \centering
    \includegraphics[width=1\linewidth]{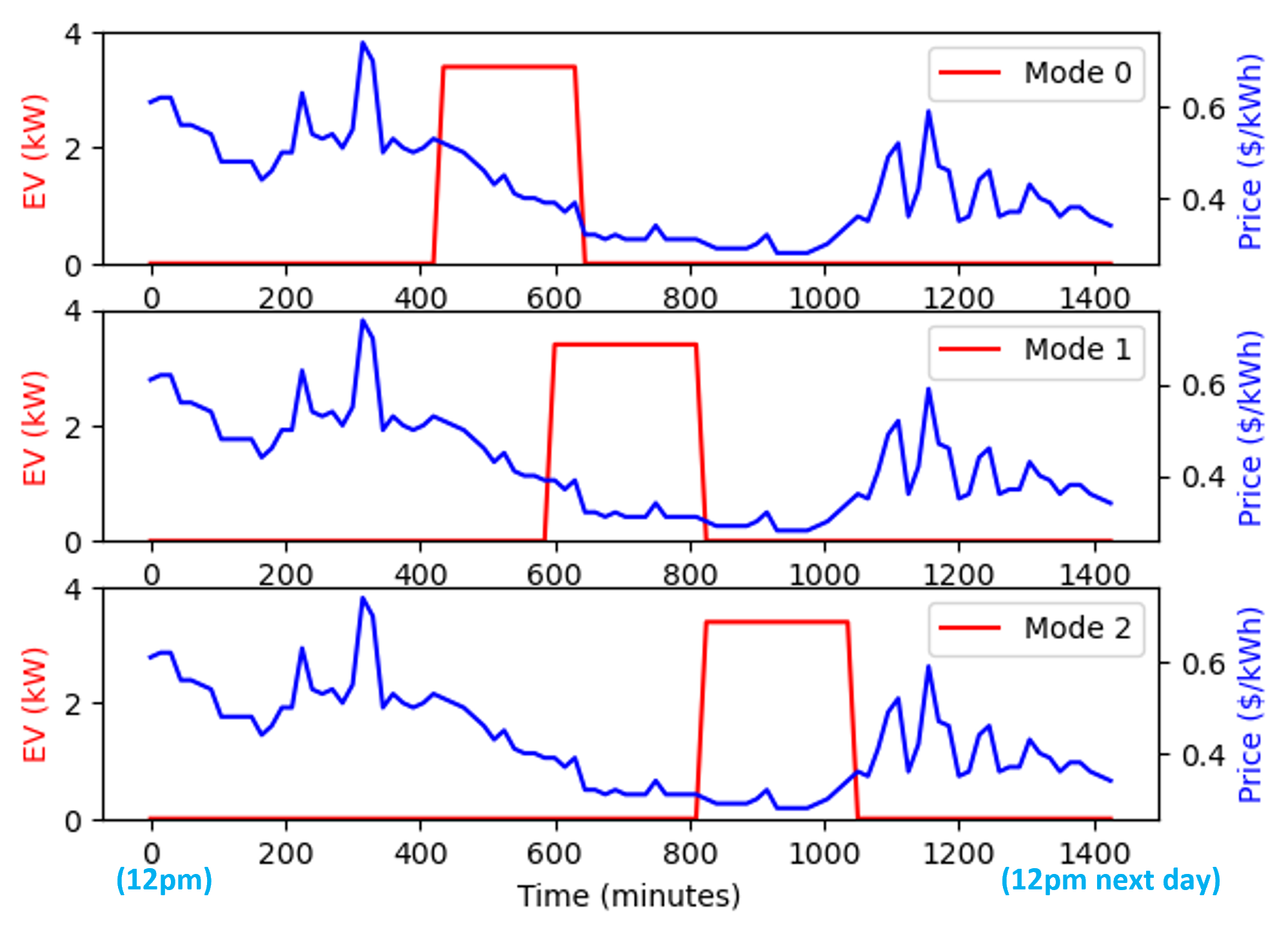}
    \caption{Optimal charging of electric vehicle with different preferences}
    \label{fig:EV}
\end{figure}

\begin{figure}
    \centering
    \includegraphics[width=1\linewidth]{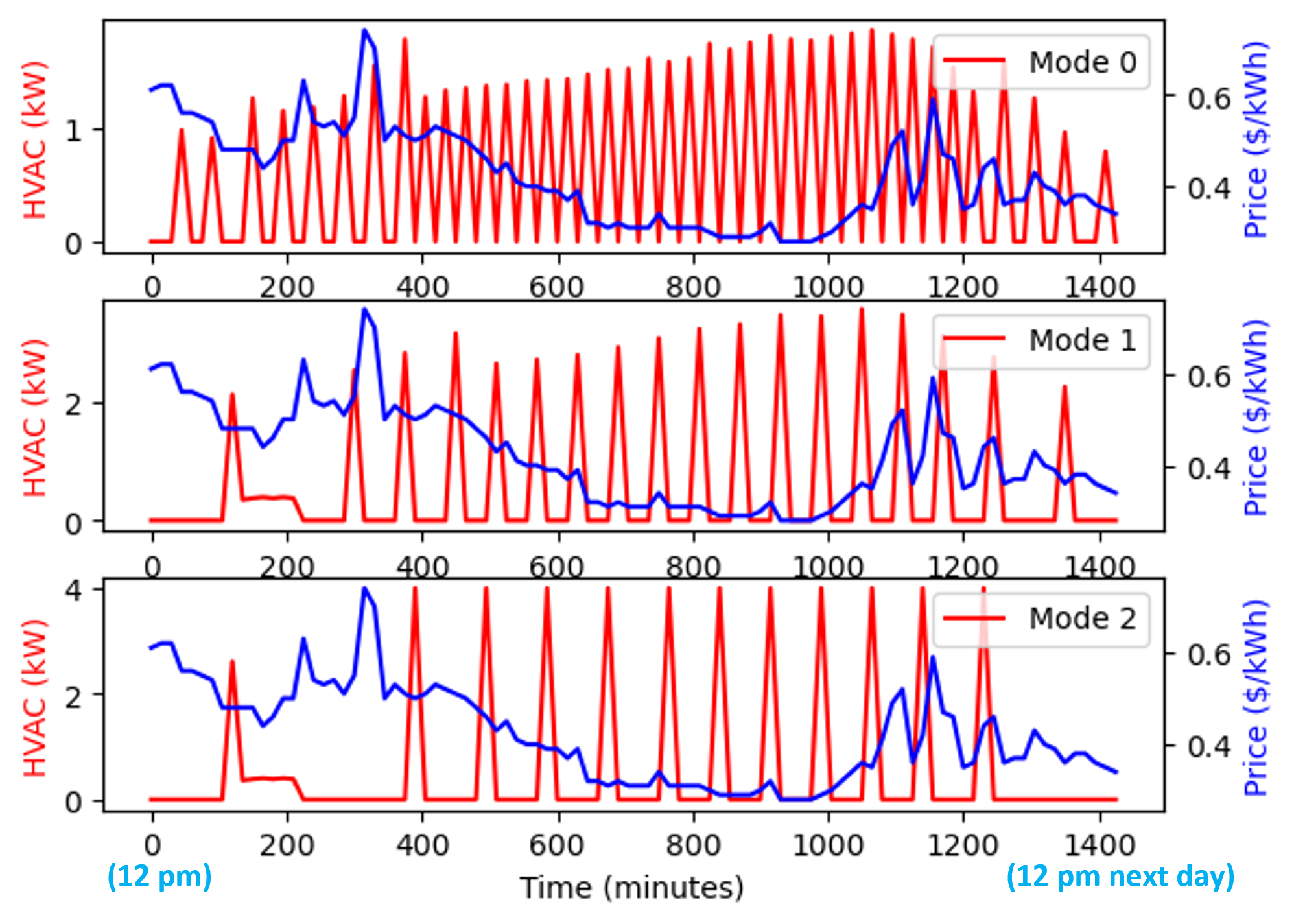}
    \caption{Power consumption of HVAC with different preferences}
    \label{fig:HVAC}
\end{figure}

\begin{figure}
    \centering
    \includegraphics[width=1\linewidth]{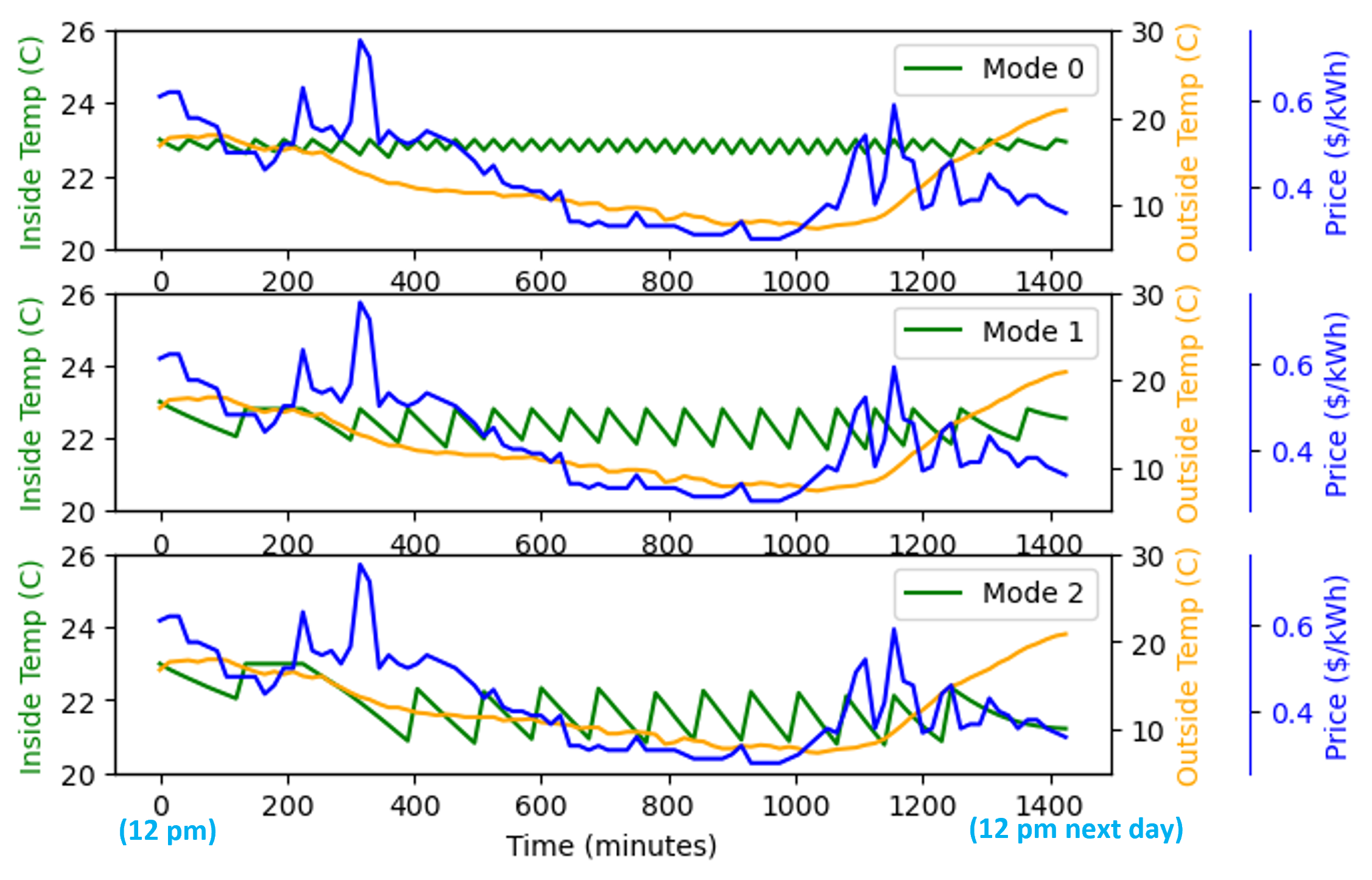}
    \caption{Outdoor and corresponding indoor temperatures for each mode}
    \label{fig:Temp}
\end{figure}

Next, we discuss the testing and evaluation results for a 24-hour period \textcolor{black}{starting from 12 pm} performed by the trained agent to analyze the optimal load management of a smart home with multi-mode preferences. Fig \ref{fig:DW},  \ref{fig:WM}, and \ref{fig:EV} illustrate how the DRL agent strategically schedules the operational timings of time-shiftable appliances and EV to align with periods of lower energy costs, reflecting the varying preferences inherent in each mode. As illustrated in Fig~\ref{fig:DW}, the WM in mode 2 commences operations earlier and at a higher price compared to mode 1, despite its more extended scheduling time interval. This behavior can be attributed to the fact that the average day-ahead price within its interval exceeds that of mode 1.
Fig \ref{fig:HVAC} details the HVAC system's power consumption, while \ref{fig:Temp} showcases the corresponding indoor and outdoor temperature dynamics.
The agent's approach to HVAC energy management is strategic, adjusting to three unique mode preferences that are characterized by varying set point temperature thresholds and modulating the HVAC demand in accordance with anticipated average energy costs. In Mode 0, the focus is on maintaining a consistent indoor temperature, demonstrating a preference for comfort over cost efficiency by largely ignoring fluctuating price signals. Modes 1 and 2, however, display a more price-sensitive pattern of energy consumption, with marked reductions in HVAC usage during times of high energy costs and a willingness to adopt a more variable indoor temperature range. This behavior accentuates the agent's ability to navigate the trade-off between energy cost savings and the provision of comfortable indoor climate conditions. The inclusion of these comfort thresholds within the state space for each appliance ensures that the trained agent possesses the capability to adjust to variations in comfort settings as they arise, showcasing a sophisticated level of adaptability.

These results affirm the robustness of our trained model. Crafted with intrinsic responsiveness, it keenly perceives and adjusts to changing environmental states, ensuring optimal energy consumption aligns with the chosen preference mode, thus bolstering both flexibility and efficiency. Table \ref{tab:cost} offers a detailed comparison of electricity costs across different modes, underscoring appreciable cost savings inherent to each mode. This financial efficacy is attributed to the model's ability to leverage the day-ahead energy prices, calculating an average prediction price that becomes instrumental in driving these savings. Thus, our model not only epitomizes adaptability but also showcases tangible benefits in terms of cost-effectiveness based on real-world energy market dynamics.

\begin{table}[htbp]
  \centering
  \caption{Electricity cost for each mode over a 24-hour period}
    \begin{tabular}{m{2cm} m{1.5cm} m{1.5cm} m{1.5cm}}
    \toprule[1.5pt]
          & \multicolumn{3}{c}{Electricity Cost (\$/day)} \\
          & Mode 0 & Mode 1 & Mode 2 \\
    \midrule[1.5pt]
    DW    & 0.43  & 0.28  & 0.27 \\
    \midrule
    WM    & 0.29  & 0.17    & 0.19 \\
    \midrule
    HVAC  & \textcolor{black}{5.91}  & \textcolor{black}{5.58}  & \textcolor{black}{4.67} \\
    \midrule
    EV    & 5.18  & 4.19  & 3.78 \\
    \midrule
    Total Cost & \textcolor{black}{11.81} & \textcolor{black}{10.22} & \textcolor{black}{8.91} \\
    \bottomrule[1.5pt]
    \end{tabular}%
  \label{tab:cost}%
\end{table}%

To evaluate the effectiveness of our model, we benchmarked it against a Mixed-Integer Linear Programming (MILP) algorithm as shown in Table \ref{tab:MILP}, utilizing the Gurobi solver with appliance preferences set to mode 2. The MILP, modeled with full knowledge of the system dynamics and future states (see the mathematical formulation in the appendix), allowing it to make globally optimal scheduling decisions. Unlike heuristic or gradient-based methods, MILP systematically evaluates all possible solutions over the entire scheduling horizon, ensuring optimal cost minimization by leveraging perfect knowledge of electricity prices, appliance demands, and operational constraints.

As a result, MILP achieves a slightly better energy cost, with an 8.69 daily cost compared to our DRL model's 8.91. This represents a difference of only 2.5$\%$, highlighting the DRL model's ability to approach near-optimal solutions based on the learned knowledge to select the actions that maximize the
reward. Furthermore, our trained agent demonstrates remarkable efficiency in terms of computational expediency, requiring merely one second to determine an optimal decision, whereas MILP demands a relatively protracted five seconds. 
\textcolor{black}{While the DRL-HEMS schedules result in an energy cost that is only 2.5$\%$ higher than the optimal achieved by MILP, this demonstrates that the DRL-HEMS is highly effective at learning cost-efficient schedules for HEMS with multi-mode preferences. Moreover, it naturally handles system uncertainties with significantly less computation time, making it a robust and practical solution for real-time energy management.}

\begin{table}[htbp]
  \centering
  \caption{Comparison of Dueling Double DQN and MILP algorithms}
    \begin{tabular}{m{1.5cm} m{1.5cm} m{2cm} m{1.5cm}}
    \toprule[1.5pt]
    Algorithm & Training time & Computation time  & Cost (\$/day) \\
    \midrule[1.5pt]
    DRL   & 240 min & 1 s   & \textcolor{black}{8.91} \\
    MILP  & -     & 5 s   & \textcolor{black}{8.69} \\
    \end{tabular}%
  \label{tab:MILP}%
\end{table}%

The scalability of our approach depends on whether homes operate independently or in a coordinated manner. Since the model is designed based on adaptive comfort, it can generalize to different comfort preferences without requiring fundamental modifications. For independent operation, transfer learning can fine-tune a pre-trained model to account for variations in thermal load and appliance characteristics, reducing the need for full retraining. In a coordinated setting, scalability can be enhanced through decentralized multi-agent reinforcement learning (MARL) or federated learning, allowing multiple HEMSs to optimize energy usage collectively while maintaining computational efficiency and privacy. These approaches ensure adaptability across diverse household settings without compromising performance.

%%%%%%%%%%%%%%%%%%%%%%%%%%%%%%%%%%%%%%%%%%%%%%%%%%%%%%%%%%%%%%%%%%%%%%
\section{Conclusion and Future Work}
HEMS stands as an instrumental component in the landscape of smart homes and the broader smart grid ecosystem, underpinning the aspirations of energy conservation, cost reduction, and user satisfaction. Despite the extensive research in this domain, a substantial gap persists, notably in capturing the dynamic and diverse nature of consumer comfort preferences within the optimization paradigm.
This study endeavors to fill this gap by introducing a multi-mode DRL algorithm-based HEMS. Our DRL-HEMS framework seamlessly integrates three distinct modes: Mode 1, which operates appliances based on their default settings without any DR activity; Mode 2 introduces DR with medium flexibility; and Mode 3 offers DR with high flexibility. Such a structure ensures a high degree of adaptability, allowing users to engage optimally with DR programs based on their comfort and energy-saving objectives. 
Rooted in the single-agent DRL algorithm, our framework offers ease of use, permitting users to switch between these modes effortlessly. Validated with real-world data sampled at 15-minute intervals, our model showcases proficiency in optimizing energy consumption based on selective preferences for each appliance across the three defined preference modes. Notably, when compared with traditional methods like MILP, \textcolor{black}{our approach closely matches the optimal energy cost while excelling in computational efficiency.}

Building upon the current work and results obtained, there are compelling directions to expand and deepen our study in the realm of HEMS. To start, we see the potential for incorporating a broader spectrum of appliances and resources, notably rooftop solar panels and energy storage systems. Such an integration will serve to enhance the versatility of the system, reflecting a more holistic energy consumption and generation profile within households. 
Additionally, real-world appliance performance variations, such as EV battery degradation and HVAC efficiency loss, can impact model accuracy over time. To address this, online learning techniques can be employed to dynamically update system parameters based on observed discrepancies. Reinforcement learning extensions and predictive maintenance strategies can further improve long-term performance by adapting control decisions to evolving efficiency trends, ensuring sustained accuracy in practical deployments.
Finally, a pertinent area of interest lies in the integration of a larger cohort of residential homes. By doing so, we aim to examine the implications of the rebound effect, particularly the challenges presented when peak load migrates from high-cost periods to those of lower pricing. Addressing and developing mechanisms to counteract such shifts is vital to maintaining grid stability and efficiency.
%%%%%%%%%%%%%%%%%%%%%%%%%%%%%%%%%%%%%%%%%%%%%%%%%%%%%%%%%%%%%%%%%%%%%%
% The bibliography is stored in an external database file
% in the BibTeX format (file_name.bib).  The bibliography is
% created by the following command and it will appear in this
% position in the document. You may, of course, create your
% own bibliography by using thebibliography environment as in
%
% \begin{thebibliography}{12}
% ...
% \bibitem{itemreference} D. E. Knudsen.
% {\em 1966 World Bnus Almanac.}
% {Permafrost Press, Novosibirsk.}
% ...
% \end{thebibliography}

% Here's where you specify the bibliography style file.
% The full file name for the bibliography style file 
% used for an ASME paper is asmems4.bst.
\bibliographystyle{asmems4}

% Here's where you specify the bibliography database file.
% The full file name of the bibliography database for this
% article is asme2e.bib. The name for your database is up
% to you.
\bibliography{asme2e}

%%%%%%%%%%%%%%%%%%%%%%%%%%%%%%%%%%%%%%%%%%%%%%%%%%%%%%%%%%%%%%%%%%%%%%
\clearpage
\appendix       %%% starting appendix
\section*{MILP-HEMS Formulation}

\label{MILP-Formulation}
\textcolor{black}{
\begin{align*}
 \textsf{Minimize: }\\&\hspace{-1cm}\sum_{t=1}^{T} \rho(t) \sum_{n=1}^{N} \left( p_{sa,n}(t) \cdot a_{sa,n}(t) + p_{EV}(t) \cdot a_{EV}(t) + p_{HVAC}(t)\right)\\
 \textsf{Subject to:}&\\
 &\sum_{t=t_{a,n}}^{t_{b, n, m}} a_{sa,n}(t) = d_n, \quad \forall n\\
 &a_{sa,n}(t) = 0, \quad \forall n, \forall t \notin [t_{a,n}, t_{b, n, m}]\\
 &\sum_{t=t_{a,n}}^{t_{b, n, m}} g_{sa,n}(t) = 1, \quad \forall n\\
 &a_{sa,n}(t+j) \geq g_{sa,n}(t), \quad \forall n, \forall t \in [t_{a,n}, t_{b, n, m}], \forall j \in [0, d_n-1]\\
 & \text{SoC}(t) = \text{SoC}(t-1) + \frac{a_{\text{EV}}(t) \cdot p_{\text{EV}} \cdot \eta_{\text{EV}} \cdot \Delta t}{C_{\text{battery}}}, \quad \forall t\\
 & SoC_{\text{min}} \leq SoC(t) \leq SoC_{\text{max}}, \quad \forall t\\
 & SoC(t_{\text{dep, m}}) = SoC_{\text{max}}\\
 & a_{\text{EV}}(t) = 0, \quad \forall t
\notin [t_{arr}, t_{dep, m}]\\
& p_{HVAC}(t) = \frac{Q_{\text{HVAC}}(t)}{\text{CoP}}, \quad \forall t\\
& Q(t) = \frac{T_{\text{set}} - T_{\text{out}}(t) + (T_{\text{out}}(t) - T_{\text{in}}(t-1)) \cdot e^{-\frac{\Delta t}{R \cdot C}}}{R \cdot (1 - e^{-\frac{\Delta t}{R \cdot C}})}, \quad \forall t \\
& Q_{\text{HVAC}}(t) \leq  Q_{\max} \cdot a_{\text{HVAC}}(t), \quad \forall t\\
& Q_{\text{HVAC}}(t) \geq  -Q_{\max} \cdot a_{\text{HVAC}}(t), \quad \forall t\\
& Q_{\text{HVAC}}(t) \geq Q(t) - (1 - a_{\text{HVAC}}(t)) \cdot M, \quad \forall t\\
& Q_{\text{HVAC}}(t) \leq Q(t) + (1 - a_{\text{HVAC}}(t)) \cdot M, \quad \forall t\\
& T_{\text{in}}(t) = T_{\text{out}}(t) + Q_{\text{HVAC}}(t) \cdot R - \left(T_{\text{out}}(t) + Q_{\text{HVAC}}(t) \cdot R - T_{\text{in}}(t-1)\right) \cdot e^{-\frac{\Delta t}{R \cdot C}}, \quad \forall t\\
& T_{\text{min}} \leq T_{\text{in}}(t) \leq T_{\text{max}}, \quad \forall t\\
\end{align*}
}

\textbf{Additional Variables and Parameters Description:}
\begin{description}
    \item[$g_{sa,n}(t) :$]{A binary indicator variable that ensures the sequential activation of the shiftable appliance $n$ over time $t$}
    \item[$M$ :]{A large constant used in the Big-M method to enforce the correct operation of the HVAC system. It guarantees that the HVAC operates at the required heat rate when the binary control variable $a_{\text{HVAC}}(t)$ is active. }
    \item[$Q_{\text{HVAC}}(t)$ :]{The actual heat rate applied to the HVAC system at time step $t$.}
\end{description}

\end{document}